\documentclass[letterpaper]{article} % DO NOT CHANGE THIS
\usepackage{aaai25}  % DO NOT CHANGE THIS
\usepackage{times}  % DO NOT CHANGE THIS
\usepackage{helvet}  % DO NOT CHANGE THIS
\usepackage{courier}  % DO NOT CHANGE THIS
\usepackage[hyphens]{url}  % DO NOT CHANGE THIS
\usepackage{graphicx}
\usepackage{float}
\usepackage{subfig}
\usepackage{natbib}  % DO NOT CHANGE THIS AND DO NOT ADD ANY OPTIONS TO IT
\usepackage{caption} % DO NOT CHANGE THIS AND DO NOT ADD ANY OPTIONS TO IT
\usepackage{algorithm}
\usepackage{algorithmic}
\usepackage{color}
\usepackage{amsmath}
\usepackage{amssymb}
\usepackage{mathrsfs}
\usepackage{multirow}

\usepackage{newfloat}
\usepackage{listings}
\DeclareCaptionStyle{ruled}{labelfont=normalfont,labelsep=colon,strut=off} % DO NOT CHANGE THIS
\floatstyle{ruled}
\newfloat{listing}{tb}{lst}{}
\floatname{listing}{Listing}
\title{Cross-Modal Spherical Aggregation for Weakly Supervised Remote Sensing Shadow Removal}
\author{
Kaichen Chi\textsuperscript{\rm 1}, Wei Jing\textsuperscript{\rm 1}, Junjie Li\textsuperscript{\rm 1}, Qiang Li\textsuperscript{\rm 1}, Qi Wang\textsuperscript{\rm 1}\thanks{Corresponding author.}\\
}
\affiliations{
    \textsuperscript{\rm 1}School of Artificial Intelligence, Optics and Electronics (iOPEN), Northwestern Polytechnical University, Xi'an, China \\
\{chikaichen, wei\_adam\}@mail.nwpu.edu.cn, \{lij55891, liqmges, crabwq\}@gmail.com
}

\usepackage{bibentry}

\begin{document}

\maketitle

\begin{abstract}
Remote sensing shadow removal, which aims to recover contaminated surface information, is tricky since shadows typically display overwhelmingly low illumination intensities.
In contrast, the infrared image is robust toward significant light changes, providing visual clues complementary to the visible image.
Nevertheless, the existing methods ignore the collaboration between heterogeneous modalities, leading to undesired quality degradation.
To fill this gap, we propose a weakly supervised shadow removal network with a spherical feature space, dubbed S2-ShadowNet, to explore the best of both worlds for visible and infrared modalities.
Specifically, we employ a modal translation (visible-to-infrared) model to learn the cross-domain mapping, thus generating realistic infrared samples.
Then, Swin Transformer is utilized to extract strong representational visible/infrared features.
Simultaneously, the extracted features are mapped to the smooth spherical manifold, which alleviates the domain shift through regularization.
Well-designed similarity loss and orthogonality loss are embedded into the spherical space, prompting the separation of private visible/infrared features and the alignment of shared visible/infrared features through constraints on both representation content and orientation.
Such a manner encourages implicit reciprocity between modalities, thus providing a novel insight into shadow removal.
Notably, ground truth is not available in practice, thus S2-ShadowNet is trained by cropping shadow and shadow-free patches from the shadow image itself, avoiding stereotypical and strict pair data acquisition.
More importantly, we contribute a large-scale weakly supervised shadow removal benchmark, including 4000 shadow images with corresponding shadow masks.
Extensive experiments demonstrate that S2-ShadowNet outperforms state-of-the-art methods in both qualitative and quantitative comparisons.
The code and benchmark are available at {\color{magenta}\url{https://github.com/chi-kaichen/S2-ShadowNet}}.
\end{abstract}


\section{Introduction}
The shadow is a prevalent physical phenomenon in nature, typically formed when light sources are occluded.
Unfortunately, undesired shadows bring further complexities and challenges to subsequent vision tasks, \textit{e.g.}, object detection, semantic segmentation, and scene classification.
Therefore, shadow removal is a nontrivial step in computer vision processing, and has received increasing attention.

\begin{figure}[t]
\centering
\includegraphics[width=0.85\columnwidth]{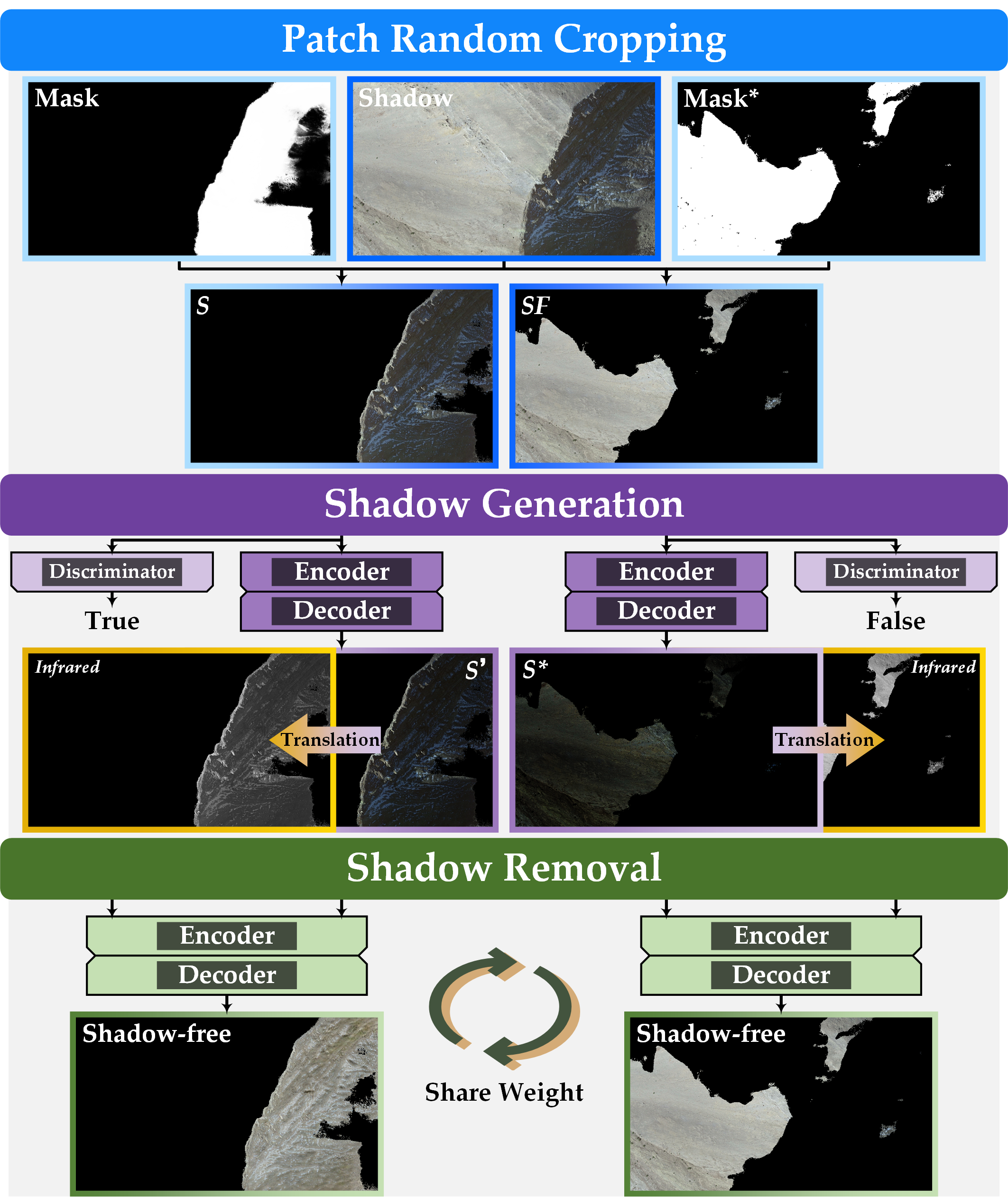}
\caption{The schematic illustration of our basic idea. S2-ShadowNet crops shadow and shadow-free regions from the image itself to control domain gap. More importantly, the introduction of complementary information from diverse modalities stimulates the potential of shadow removal.}
\label{fig1}
\end{figure}
Early shadow removal methods analyze the statistics of illumination to detect and remove shadows through prior knowledge, such as gradient \cite{Finlayson:TPAMI-06}, morphology \cite{Silva:ISPRS-18}, illumination condition \cite{Finlayson:IJCV-09}, and user interaction \cite{Gong:BMVC-14}.
However, the handcrafted prior is fundamentally ill-posed, leading to traditional methods are unstable and sensitive when facing challenging shadow distortions.

Benefiting from the remarkable generalization capability of neural networks, deep learning-based methods lead to substantial improvement in shadow removal.
Pioneering works have been explored from diverse perspectives, \textit{e.g.}, multi-task decoupling \cite{Liu:TMM-23} and exposure fusion \cite{Fu:CVPR-21}.
However, the above methods employ paired shadow and shadow-free versions to train neural networks in a fully supervised manner.
As is well known, capturing high-quality image pairs in uncontrolled natural environments is both difficult and expensive.
To this end, recent researches \cite{Hu:ICCV-19, Jin:ICCV-21, Liu:TIP-21} concentrate on unsupervised learning to shadow removal.
Unfortunately, unsupervised methods typically produce inconsistent hue and luminosity due to the huge domain gap between non-corresponding shadow and shadow-free images \cite{Le:ECCV-20, Liu:CVPR-21}.
Therefore, weakly supervised learning for cropping content-similar shadow and shadow-free regions from the same image is preferable.
Such a manner is attractive in controlling domain discrepancies.

In this paper, S2-ShadowNet crops region pairs based on masking operation, coupled with a shadow generation strategy that absorbs the desirable property of dark pixels to generate pseudo shadows for shadow-free regions, thus constructing training signals that support weakly supervised learning, as depicted in Figure \ref{fig1}.
Subsequently, the shadow removal paradigm accomplishes revolutionary shadow recovery effects through cross-modal feature decomposition, beyond that of a single modality.
Specifically, visible and infrared features are projected into a spherical space, and then under the dual constraints of orthogonality and similarity, illumination components shared between modalities are aligned while unique private features (texture and thermal information, \textit{etc.}) are inversely separated.
Without bells and whistles, the dependencies between diverse modalities are gracefully modeled.
In a nutshell, the main contributions can be summarized in three-fold:
\begin{itemize}
\item \textbf{Perspective contribution.}
We rethink the shadow removal task from the perspective of multi-modal collaboration, relighting shadow pixels through the dynamic update of illumination-related dominant modality.
To our best knowledge, this is the first attempt to remove shadow effects by introducing infrared knowledge.

\item \textbf{Technical contribution.}
We propose a cross-modal spherical aggregation network to explore the interaction of visible and infrared modalities.
Multi-modal feature representations are mapped to the spherical space, and the inner product between private and shared vectors is employed to drive separation and alignment, thus removing shadow traces.

\item \textbf{Practical contribution.}
We construct a large-scale real-world weakly supervised shadow removal benchmark (WSSR), which provides a platform for qualitative to quantitative analysis and motivates the development of deep learning-based shadow removal.
\end{itemize}

\section{Related Work}
In this section, we first review multi-modal image translation and multi-modal image collaboration strategies, followed by a discussion on existing shadow removal methods.

\noindent
\textbf{Multi-modal Image Translation.}
In view of the robustness of infrared imaging in poor illumination environments, visible-to-infrared translation schemes have been actively investigated.
Zhang \textit{et al.} \cite{Zhang:TIP-19} employed a pix2pix-based translation component and explored the potential capability of frequency filters to alleviate data differ.
Similarly, Li \textit{et al.} \cite{Li:TNNLS-21} translated cityscapes under diverse lighting conditions to thermal cityscapes based on pix2pixHD, thus serving the subsequent semantic urban scenario understanding.
Devaguptapu \textit{et al.} \cite{Devaguptapu:CVPRW-19} designed a visible semantics-guided pseudo multi-modal translation framework, which borrows domain adaptation knowledge for cross-modal alignment.
In summary, the booming development of multi-modal image translation technology alleviates the scarcity of infrared data, making possible multi-modal collaboration.

\noindent
\textbf{Multi-modal Image Collaboration.}
Multi-modal image collaboration aims to enjoy the mutual benefits between multiple sensors, thus achieving superior performance on practical tasks.
Cao \textit{et al.} \cite{Cao:ICCV-23} contributed a multi-modal gated mixture framework, which recovers texture and contrast in low-light scenarios by integrating local experts and global experts.
Zhang \textit{et al.} \cite{Zhang:CVPR-23} embedded a center-guided visible-infrared pair mining loss into cross-modality networks to alleviate the interference of significant light changes for person re-identification.
Xie \textit{et al.} \cite{Xie:TITS-24} proposed a hallucination network to integrate visible and thermal infrared information, and then used illumination-aware loss to regress visible features from similar hallucination. Such a manner effectively promotes the robustness of nighttime pedestrian detection.
Obviously, the complementarity of multiple modalities is a quite promising direction, deserving to be studied in the context of shadow removal.

\noindent
\textbf{Shadow Removal.}
Traditional-based methods typically focus on exploring diverse physical shadow properties.
Guo \textit{et al.} \cite{Guo:TPAMI-13} predicted relative illumination conditions between shadow and shadow-free regions to remove shadow effects.
Gong \textit{et al.} \cite{Gong:BMVC-14} employed user interaction of shadow and lit regions to highlight shadow boundary intensity changes, and then accurately and robustly removed shadows by inverse scaling non-uniform field.
Unfortunately, traditional-based methods hardly cover real-world shadow detection and removal due to the high dependence on specific physical priors.

\begin{figure*}[t]
\centering
\includegraphics[width=0.95\textwidth]{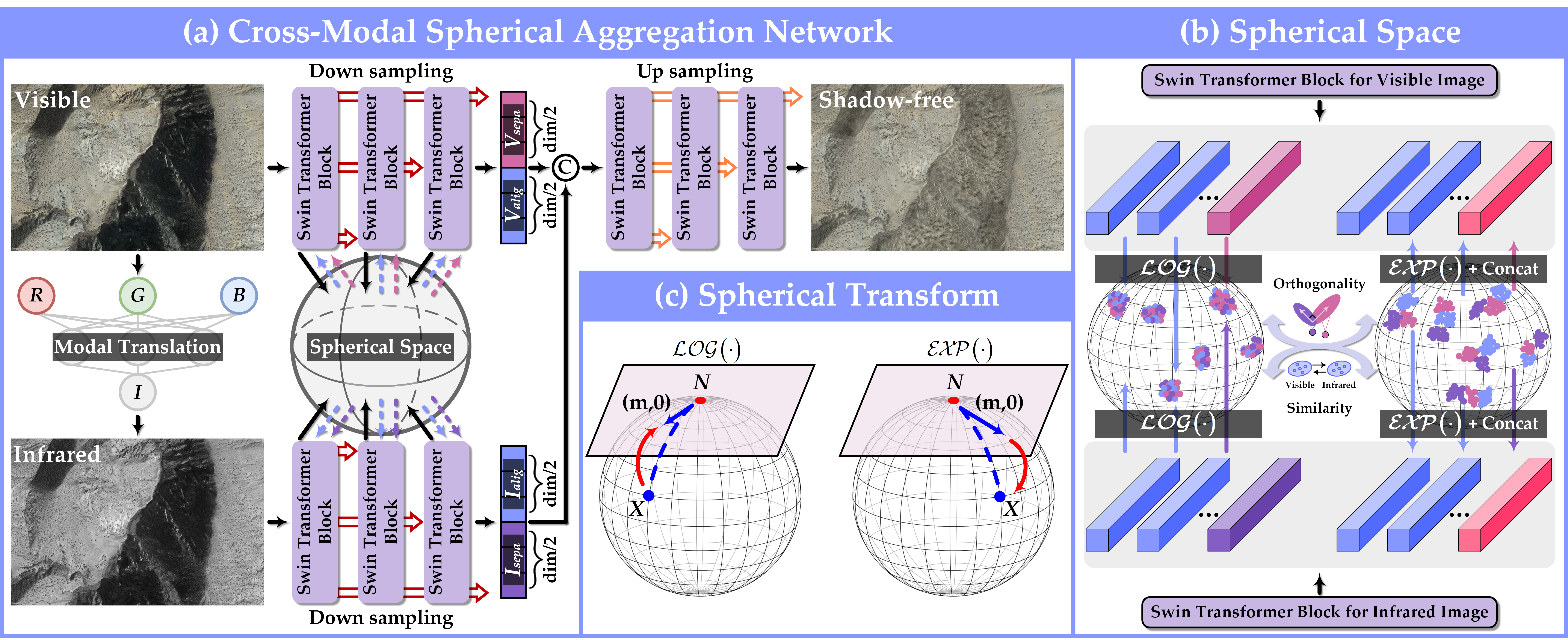}
\caption{(a) Model Architecture. We employ pretrained U2Fusion to generate infrared samples. Multi-modal representations are projected into the spherical space through spherical transform to accomplish decomposition, \textit{i.e.}, alignment and separation. Under both spatial-wise and pixel-wise constraints, the full visible feature and the shared infrared feature are integrated into a shadow-free image. (b) The schematic illustration of spherical space. (c) The schematic illustration of spherical transform.}
\label{fig2}
\end{figure*}
Most recently methods are based on deep learning
Wan \textit{et al.} \cite{Wan:ECCV-22} employed a style estimator to explore the style representation of shadow-free regions, and then designed learnable normalization to accomplish harmonious visual appearance.
Zhu \textit{et al.} \cite{Zhu:CVPR-22} coupled the learning processes of shadow removal and shadow synthesis in a unified framework, and then recovered colors and background contents through two-way constraints.
Works \cite{Le:ECCV-20, Liu:CVPR-21} closely related to this paper cropped windows or random regions from the same image to learn mappings between shadow and shadow-free domains.
Nevertheless, these shadow removal methods neglect statistical co-occurrences between non-consistent modalities, leading to unsatisfactory robustness and generalization.

\section{Methodology}
The key components of S2-ShadowNet include a modal translation stage, a shadow generation stage, and a shadow removal stage.
The modal translation stage employs pretrained U2Fusion \cite{Xu:TPAMI-22} to provide realistic infrared signals, while the shadow generation stage guides shadow appearance rendering through adversarial learning, and we refer the readers to \cite{Liu:CVPR-21} for more details.
As a unique technical contribution, the shadow removal stage utilizes spherical space to integrate visible path and infrared path, while the joint loss constraint serves as a tool to drive the alignment and separation of multi-modal representations, as shown in Figure \ref{fig2}.
In what follows, we detail the shadow removal stage, \textit{i.e.}, the cross-modal spherical aggregation network.

\noindent
\textbf{Spherical Transform.}
Since the distances between spherical representations are regularized, multi-modal representations with various scales and orders of magnitude can be favorably aligned and separated without losing their respective unique properties \cite{Zhao:ICCV-23}.
Therefore, spherical feature learning has advantages in domain adaptation \cite{Gu:ICCV-20}.
Specifically, the spherical space transform is composed of an exponential mapping, a tangent space transform, and a logarithmic mapping \cite{Gu:TPAMI-24}, which can be expressed as:
\begin{equation}
\label{eq1}
\mathcal{T} =  \mathcal{EXP}_{N}(\mathcal{L}(\mathcal{LOG}_{N}(\mathcal{M}))),
\end{equation}
where $\mathcal{EXP}(\cdot)$, $\mathcal{L}(\cdot)$, and $\mathcal{LOG}(\cdot)$ represent the spherical exponential mapping, the linear transform in tangent space, and the spherical logarithmic mapping.
Besides, $\mathcal{M}$ represents multi-modal features, $N = (0,...,0,r)$ represents the north pole, and $r$ represents the radius of spherical space.

\begin{figure}[t]
\centering
\includegraphics[width=0.8\columnwidth]{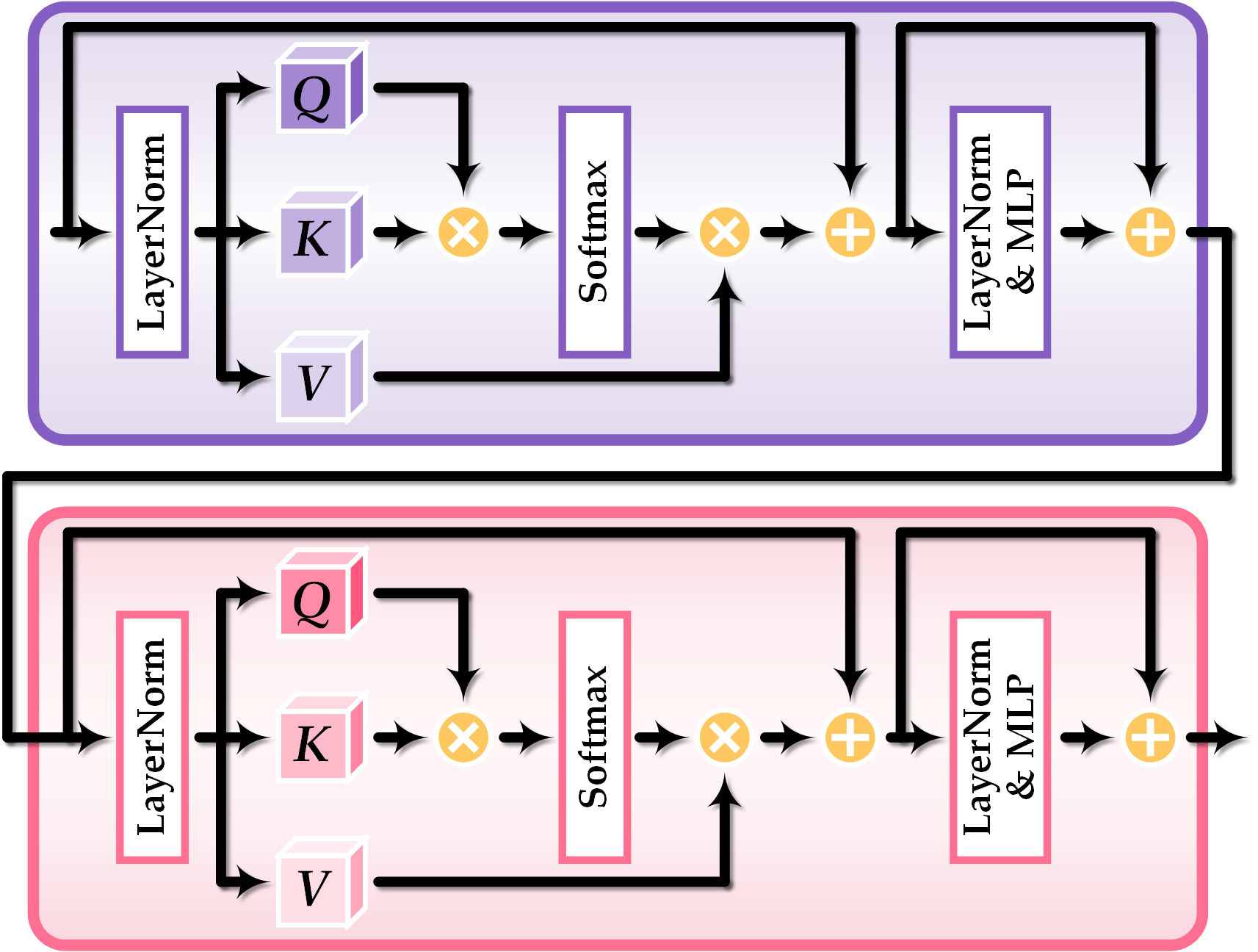}
\caption{The architecture of the Swin Transformer block.}
\label{fig3}
\end{figure}
As depicted in Figure \ref{fig2}(c), the exponential and logarithmic mappings connect the tangent space $T_{N}$ and the spherical space $\mathbb{S}_{r}^{n-1} = \left\{X \in \mathbb{R}^{n}:\Vert X \Vert = r\right\}$.
Given a vector $\hat{m} = (m,0)$ in $T_{N}$ starting at $N$, $\Vert \hat{m} \Vert$ on $T_{N}$ is the same as the geodesic distance between $N$ and $X$ on $\mathbb{S}_{r}^{n-1}$ \cite{Shang:TGRS-23}.
Based on this, the spherical logarithmic mapping $\mathcal{LOG}_{N}: \mathbb{S}_{r}^{n_{1}-1} \rightarrow T_{N_{1}}$ can be expressed as:
\begin{equation}
\label{eq2}
\begin{split}
\mathcal{LOG}_{N}(X) =& \frac{\alpha}{\sin \alpha}(X-N\cos\alpha), \quad \forall X \in \mathbb{S}_{r}^{n_{1}-1}\\
&\alpha = \arccos(N^{T}X/r^{2}),
\end{split}
\end{equation}

The linear transform achieves the feature propagation $T_{N_{1}} \rightarrow T_{N_{2}}$ between tangent spaces:
\begin{equation}
\label{eq3}
\mathcal{L}(\hat{m},r) =  (Wm+b,r), \quad \forall \hat{m} \in T_{N_{1}}
\end{equation}
where $W\in\mathbb{R}^{(n_{2}-1)*(n_{1}-1)}$ and $b\in\mathbb{R}^{n_{2}-1}$.
The spherical exponential mapping $\mathcal{EXP}_{N}: T_{N_{2}} \rightarrow \mathbb{S}_{r}^{n_{2}-1}$ can be expressed as:
\begin{equation}
\label{eq4}
\begin{split}
\mathcal{EXP}_{N}(\hat{m}) =& N\cos\beta+\hat{m}\frac{\sin\beta}{\beta}, \quad \forall \hat{m} \in T_{N_{2}}\\
&\beta = \Vert\hat{m}\Vert/r,
\end{split}
\end{equation}

\noindent
\textbf{Encoder–Decoder.}
The encoder employs cascaded Swin Transformer blocks \cite{Liu:ICCV-21} to simultaneously explore the global dependence of visible features $\mathcal{V}$ and infrared features $\mathcal{I}$.
Swin Transformer is known for the long-range context exploitation capability of the multi-head self-attention mechanism, focusing on the capture of shared and private features in the channel dimension \cite{Zhao:ICCV-23}, as shown in Figure \ref{fig3}.
Taking visible features as an example, we assume that the former $\frac{dim}{2}$ channels are shared, representing cross-modal information.
Obviously, the latter $\frac{dim}{2}$ channels are private, representing the texture of the visible object surface.
We expect to leverage the infrared modality to provide illumination representations missing in the visible modality, thus domain-shared and domain-private features should be separated and aligned, respectively.
To this end, we employ the logarithmic mapping $\mathcal{LOG}(\cdot)$ to project multi-modal features into the spherical space, accomplish feature decomposition under the supervision of orthogonality and similarity losses, and reproduce the promising illumination through the exponential mapping $\mathcal{EXP}(\cdot)$:
\begin{equation}
\label{eq5}
\begin{split}
&\psi_{\mathcal{M}}^{align} = \mathcal{S}(\mathcal{M})[0:\frac{dim}{2}],\\
&\psi_{\mathcal{M}}^{separ} = \mathcal{S}(\mathcal{M})[\frac{dim}{2}:dim],\\
&\hat{\psi}_{\mathcal{M}}^{align} = \mathcal{T}(\psi_{\mathcal{M}}^{align}),\\
&\hat{\psi}_{\mathcal{M}}^{separ} = \mathcal{T}(\psi_{\mathcal{M}}^{separ}),\\
&\hat{\psi}_{\mathcal{M}} = \mathcal{C}(\hat{\psi}_{\mathcal{M}}^{align}, \hat{\psi}_{\mathcal{M}}^{separ}),
\end{split}
\end{equation}
where $\mathcal{S}(\cdot)$ represents the Swin Transformer block and $\mathcal{C}(\cdot)$ represents the dimension concatenation operator.

For shadow removal, the full visible feature $\hat{\psi}_{\mathcal{V}}$ with rich edges and the shared infrared feature $\hat{\psi}_{\mathcal{I}}^{align}$ with robust illumination representations are beneficial.
Therefore, we integrate $\hat{\psi}_{\mathcal{V}}$ and $\hat{\psi}_{\mathcal{I}}^{align}$ in the channel dimension to accomplish shadow contamination removal:
\begin{equation}
\label{eq6}
\mathcal{S}_{f} = \mathcal{S}(\mathcal{C}(\hat{\psi}_{\mathcal{V}},\hat{\psi}_{\mathcal{I}}[0:\frac{dim}{2}])).
\end{equation}
where $\mathcal{S}_{f}$ represents the shadow-free image.

\begin{figure}[t]
\centering
\includegraphics[width=0.95\columnwidth]{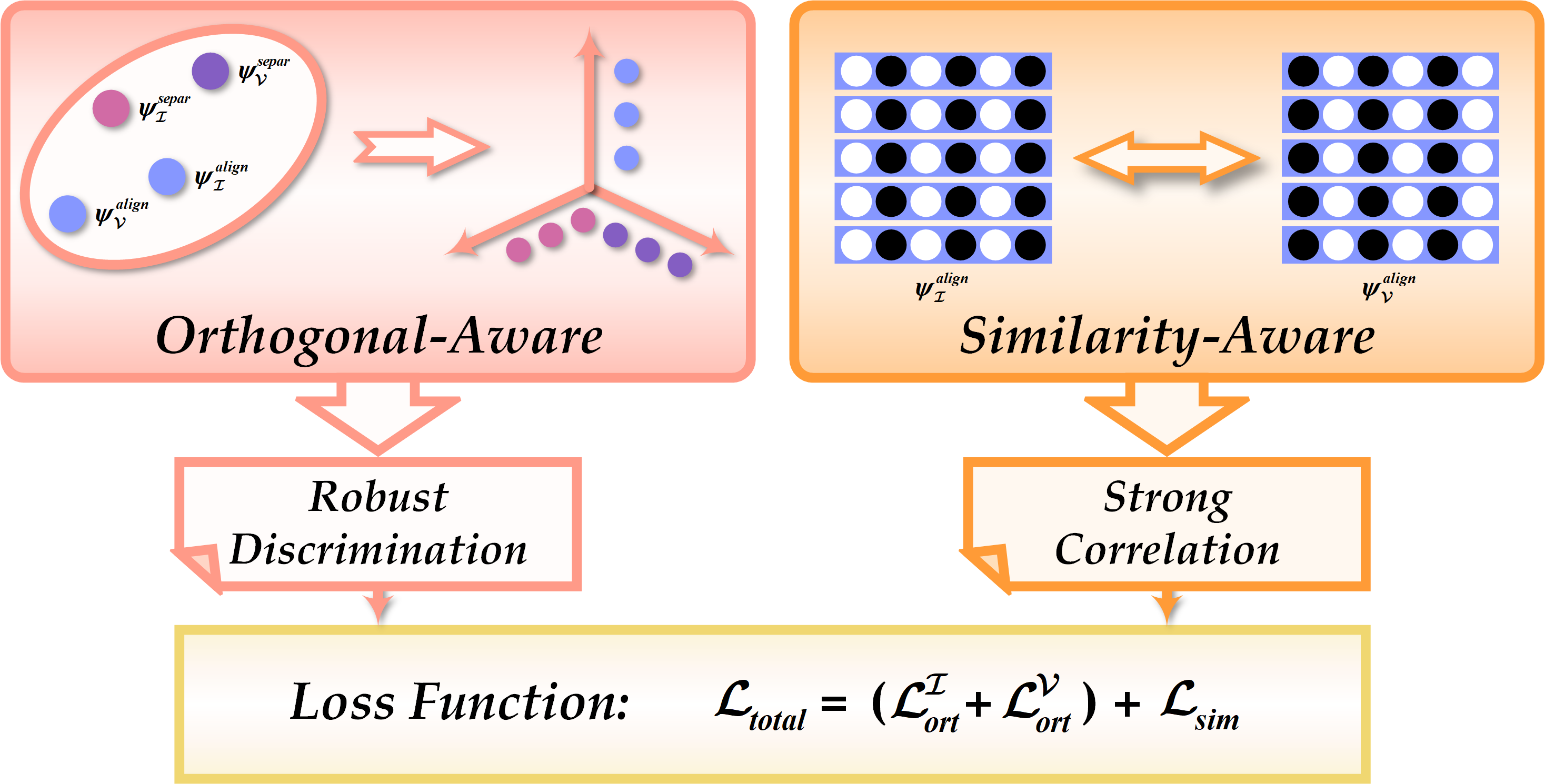}
\caption{The pipeline of the loss function in the shadow removal stage. The orthogonality loss separates independence by calculating the inner product between shared and private features. The similarity loss builds a bridge to explore semantic similarities and differences between shared features.}
\label{fig4}
\end{figure}
\noindent
\textbf{Loss Function.}
In order to achieve the decomposition of shared and private properties in a weakly supervised manner, we employ a linear combination of the orthogonality loss $\mathcal{L}_{ort}$, the similarity loss $\mathcal{L}_{sim}$, the adversarial loss $\mathcal{L}_{adv}$, and the identical loss $\mathcal{L}_{ide}$, which can be expressed as:
\begin{equation}
\label{eq7}
\begin{split}
&\mathcal{L}_{total} = \mathcal{L}_{ort}^{\mathcal{I}}(\psi_{\mathcal{I}}^{align},\psi_{\mathcal{I}}^{separ}) + \mathcal{L}_{ort}^{\mathcal{V}}(\psi_{\mathcal{V}}^{align},\psi_{\mathcal{V}}^{separ})\\
&+ \mathcal{L}_{sim}(\psi_{\mathcal{I}}^{align},\psi_{\mathcal{V}}^{align}) + \mathcal{L}_{adv}(G,D) + \mathcal{L}_{ide}(G),
\end{split}
\end{equation}

\begin{figure*}[t]
\centering
\includegraphics[width=0.85\textwidth]{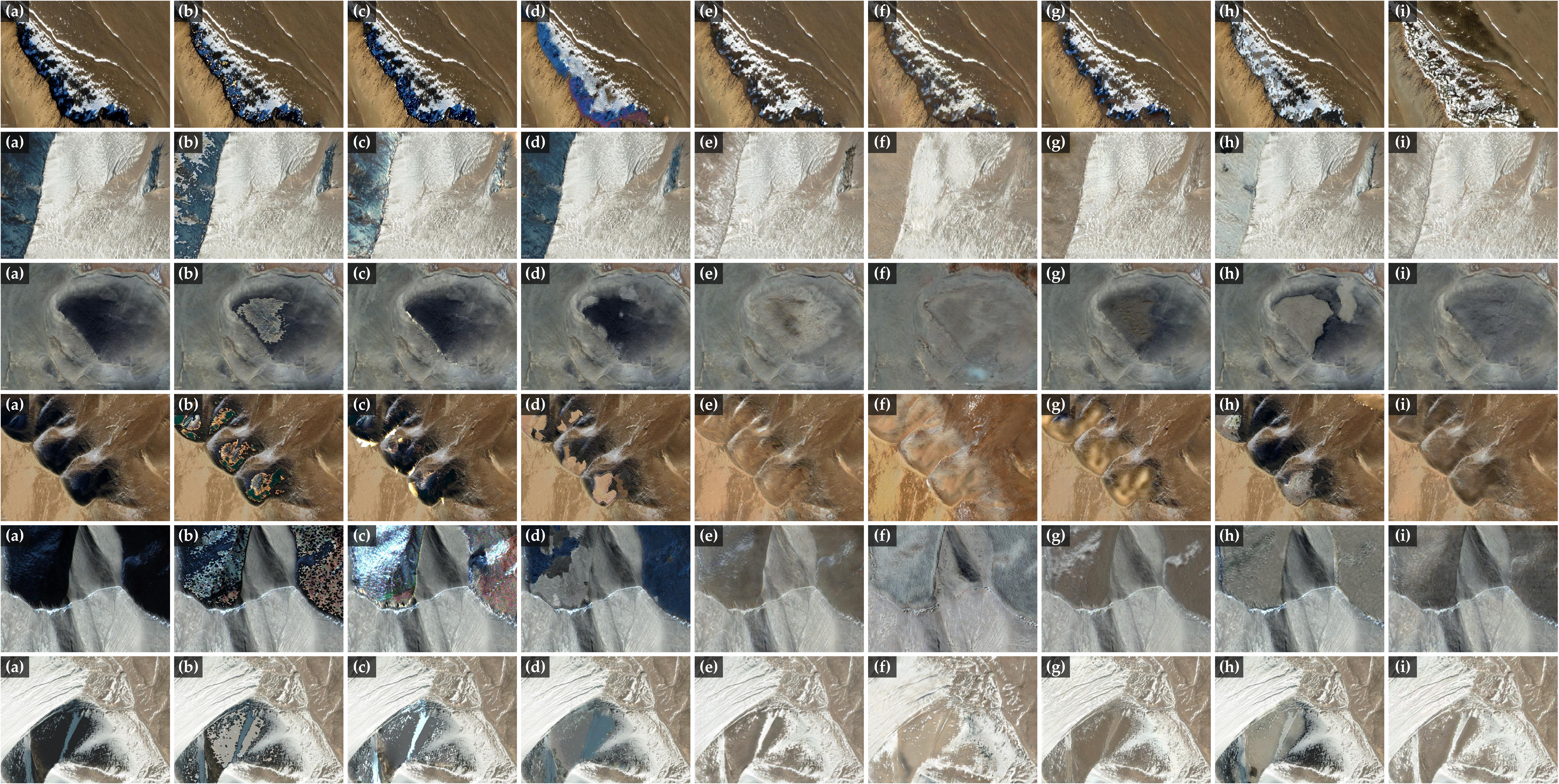}
\caption{Visual comparisons on shadow images sampled from \textbf{WSSR}. (a) input. (b) Silva. (c) Gong. (d) Guo. (e) Mask-ShadowGAN. (f) DC-ShadowNet. (g) LG-ShadowNet. (h) G2R-ShadowNet. (i) S2-ShadowNet.}
\label{fig5}
\end{figure*}
\begin{figure*}[t]
\centering
\includegraphics[width=0.85\textwidth]{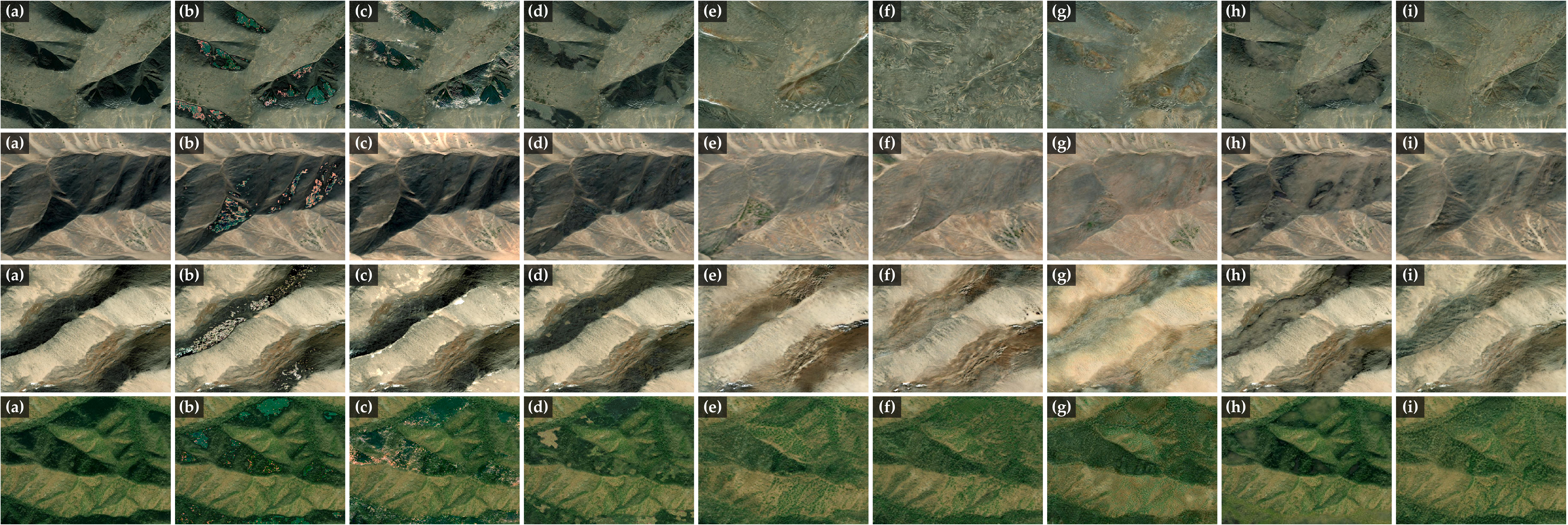}
\caption{Visual comparisons on shadow images sampled from \textbf{URSSR}. (a) input. (b) Silva. (c) Gong. (d) Guo. (e) Mask-ShadowGAN. (f) DC-ShadowNet. (g) LG-ShadowNet. (h) G2R-ShadowNet. (i) S2-ShadowNet.}
\label{fig6}
\end{figure*}
In Figure \ref{fig4}, the orthogonality loss learns an orthogonal representation from the trend of change in angle, where two distinct properties are effectively separated, displaying ideally independent distributions.
Specifically, we calculate the Gram matrices of shared and private features, and then straighten them to 1-D vectors.
The orthogonality loss is defined as the inner product between vectors:
\begin{equation}
\label{eq8}
\begin{split}
&\mathcal{L}_{ort}^{\mathcal{I}} = \mathcal{F}(\mathcal{G}(\psi_{\mathcal{I}}^{align})) \cdot \mathcal{F}(\mathcal{G}(\psi_{\mathcal{I}}^{separ})),\\
&\mathcal{L}_{ort}^{\mathcal{V}} = \mathcal{F}(\mathcal{G}(\psi_{\mathcal{V}}^{align})) \cdot \mathcal{F}(\mathcal{G}(\psi_{\mathcal{V}}^{separ})),
\end{split}
\end{equation}
where $\mathcal{F}$ represents the flattening operation and $\mathcal{G}$ represents the Gram matrix.
The similarity loss promotes consistency in illumination representations of shared features:
\begin{equation}
\label{eq9}
\begin{split}
\mathrm{SSIM}& = \frac{(2\mu_{\mathcal{I}}\mu_{\mathcal{V}}+c_{1})(2\sigma_{\mathcal{I}\mathcal{V}}+c_{2})}{(\mu_{\mathcal{I}}^{2}+\mu_{\mathcal{V}}^{2}+c_{1})(\sigma_{\mathcal{I}}^{2}+\sigma_{\mathcal{V}}^{2}+c_{2})},\\
&\mathcal{L}_{sim} = 1 - \mathrm{SSIM}(\psi_{\mathcal{I}}^{align},\psi_{\mathcal{V}}^{align}),
\end{split}
\end{equation}
where $\mu_{\mathcal{I}}$ and $\mu_{\mathcal{V}}$ represent the means of $\psi_{\mathcal{I}}^{align}$ and $\psi_{\mathcal{V}}^{align}$, respectively.
$\sigma_{\mathcal{I}}^{2}$ and $\sigma_{\mathcal{V}}^{2}$ represent the corresponding variances.
$\sigma_{\mathcal{I}\mathcal{V}}$ represents the covariance of $\psi_{\mathcal{I}}^{align}$ and $\psi_{\mathcal{V}}^{align}$.
The constants $c_{1}$ and $c_{2}$ are set to $0.01^{2}$ and $0.03^{2}$, avoiding numerical instability as the denominator approaches zero.

The adversarial loss promotes the shadow generation stage to learn the shadow distribution, while the identical loss encourages contamination generation consistent with the cropped shadow $S$:
\begin{equation}
\label{eq10}
\begin{split}
\mathcal{L}_{adv} = \mathbb{E}_{\mathcal{V},\mathcal{S}_{f}}&[\log (1-D(G(\mathcal{V})))] + \mathbb{E}_{\mathcal{V},\mathcal{S}_{f}}[\log D(\mathcal{S}_{f})],\\
&\mathcal{L}_{ide} = \mathbb{E}_{\mathcal{V}\sim p(\mathcal{V})}[\Vert G(\mathcal{V}),S\Vert_{1}].
\end{split}
\end{equation}

\section{Experiment}
\subsection{Experimental Settings}
\textbf{Implementation Details.}
The implementation of S2-ShadowNet is done with PyTorch on an NVIDIA RTX 3090 GPU.
A batch-mode learning strategy with a batch size of 2 is employed.
ADAM is applied for network optimization and the learning rate is fixed to $1e^{-4}$.

\begin{figure*}[t]
\centering
\includegraphics[width=0.85\textwidth]{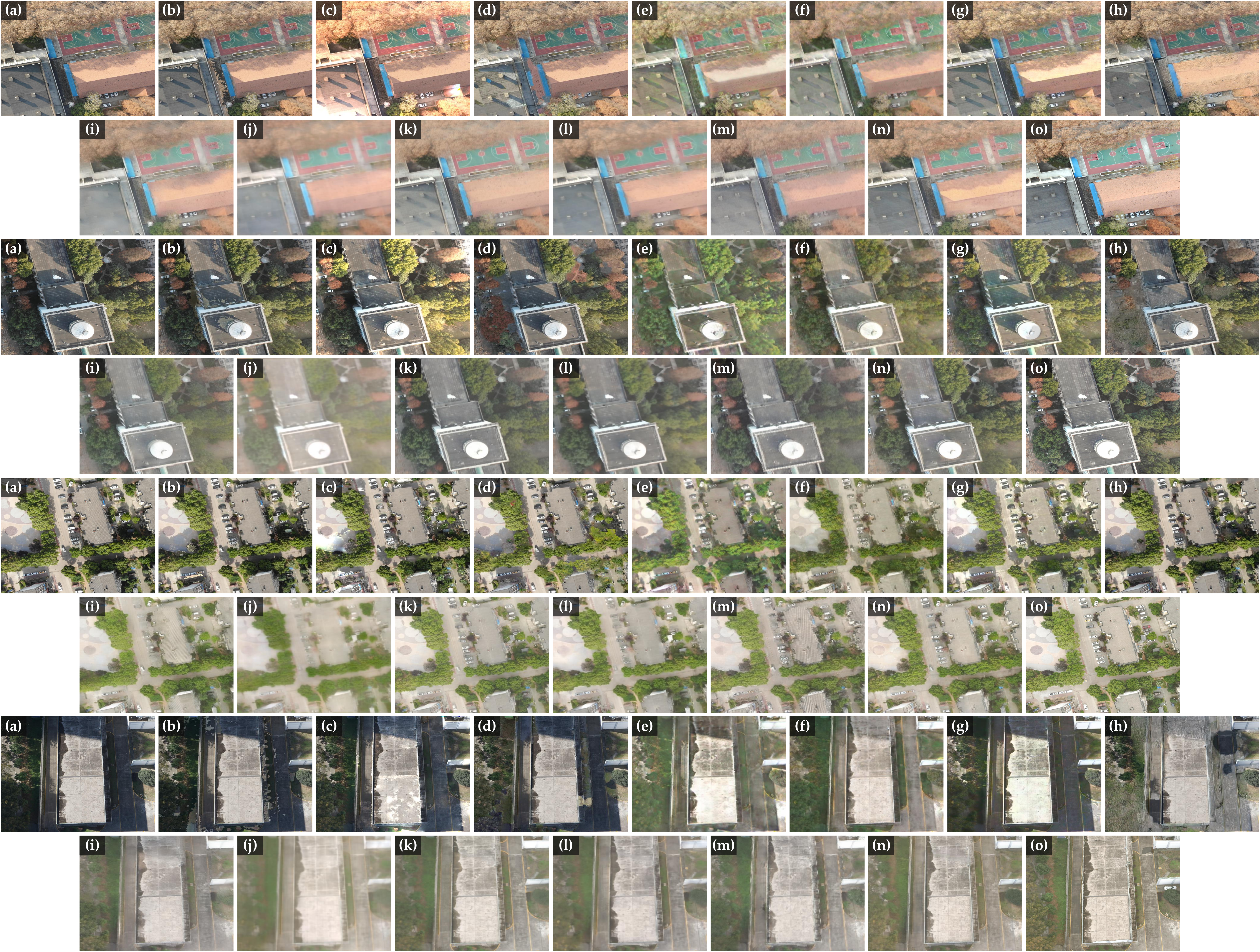}
\caption{Visual comparisons on shadow images sampled from \textbf{UAV-SC}. (a) input. (b) Silva. (c) Gong. (d) Guo. (e) Mask-ShadowGAN. (f) DC-ShadowNet. (g) LG-ShadowNet. (h) G2R-ShadowNet. (i) DMTN. (j) ST-CGAN. (k) DHAN. (l) ShadowFormer. (m) TBRNet. (n) S2-ShadowNet. (o) ground truth.}
\label{fig7}
\end{figure*}
\noindent
\textbf{Benchmarks.}
We perform qualitative and quantitative comparisons on WSSR, URSSR \cite{Wang:TGRS-24}, and UAV-SC \cite{Luo:TGRS-23} benchmarks.
\textbf{WSSR} consists of 4000 shadow images with 4K resolution from the Rocky, Appalachian, and Caucasus, \textit{etc}.
Notably, to enable weakly supervised and unsupervised learning, we also provide corresponding shadow masks \cite{Zhu:ECCV-18} and non-consistent shadow-free samples.
This further enhances the utility of WSSR.
We randomly select 3900 shadow images and corresponding shadow masks for training, while the rest 100 shadow images for testing.
\textbf{URSSR} contains unpaired 350 shadow and 230 shadow-free images captured from multiple national parks.
Similarly, we randomly select 300 shadow images while employing all shadow-free images to train S2-ShadowNet, and the rest 50 shadow images for testing.
\textbf{UAV-SC} consists of 6954 shadow images and reference ground truths, where 6924 image pairs are used for training and the rest 30 image pairs are used for testing.
UAV-SC is collected from Wuhan through two drone platforms.
In summary, the training and testing sets cover diverse mountain, woodland, and urban scenarios, different degradation characteristics, and a broad range of image content.

\noindent
\textbf{Compared Methods.}
We compare S2-ShadowNet with a series of representative shadow removal methods, including \textbf{traditional-based methods} (Silva \cite{Silva:ISPRS-18}, Gong \cite{Gong:BMVC-14}, and Guo \cite{Guo:TPAMI-13}) and \textbf{deep learning-based methods} (Mask-ShadowGAN \cite{Hu:ICCV-19}, DC-ShadowNet \cite{Jin:ICCV-21}, LG-ShadowNet \cite{Liu:TIP-21}, G2R-ShadowNet \cite{Le:ECCV-20}, DMTN \cite{Liu:TMM-23}, ST-CGAN \cite{Wang:CVPR-18}, DHAN \cite{Cun:AAAI-20}, ShadowFormer \cite{Guo:AAAI-23}, and TBRNet \cite{Liu:TNNLS-24}).

\noindent
\textbf{Evaluation Metrics.}
For UAV-SC, we perform full-reference evaluations by computing the root mean square error (\textbf{RMSE}), peak signal-to-noise ratio (\textbf{PSNR}), and structural similarity (\textbf{SSIM}).
A lower RMSE score suggests better performance, while the opposite is true for PSNR and SSIM scores.
Notably, the full-reference evaluation is computed in the shadow region (S.), non-shadow region (N.S.), and whole image (All), respectively.
For WSSR and URSSR without ground truths, we employ visual neuron matrix (\textbf{VNM}) \cite{Chang:SPL-21} and \textbf{Entropy} to perform no-reference evaluations.
A higher VNM or Entropy score indicates a more attractive visual perception.

\begin{table*}[ht]
\centering
\renewcommand\arraystretch{1.0}
\label{table1}
\setlength{\tabcolsep}{6pt}
\resizebox{!}{0.15\textwidth}{
\begin{tabular}{|c|c|c|c|c|c|c|c|c|c|}
\hline
\multirow{2}{*}{Methods} & \multicolumn{3}{c|}{RMSE($\downarrow$)} & \multicolumn{3}{c|}{PSNR($\uparrow$)} & \multicolumn{3}{c|}{SSIM($\uparrow$)}\\
\cline{2-10}
& S. & N.S. & All & S. & N.S. & All & S. & N.S. & All\\
\hline
\hline
Silva (ISPRS'18) & 32.0696 & 18.5089 & 21.3024 & 22.8902 & 18.1034 & 16.2621 & 0.8899 & 0.8035 & 0.6898\\
Gong (BMVC'14) & 26.0850 & 18.8162 & 20.3155 & 25.1448 & 18.1944 & 16.8290 & 0.9179 & 0.7926 & 0.6984\\
Guo (TPAMI'13) & 29.8381 & 17.9412 & 20.3919 & 23.5991 & 18.3971 & 16.6687 & 0.9042 & 0.8093 & 0.6945\\
Mask-ShadowGAN (ICCV'19)$\dagger$ & 19.7786 & 17.5036 & 17.9722 & 27.5729 & 20.4507 & 19.1357 & 0.9450 & 0.8142 & 0.7374\\
DC-ShadowNet (ICCV'21)$\dagger$ & 16.1495 & 13.6091 & 14.1324 & 29.2519 & 22.4207 & 21.0712 & 0.9543 & 0.8479 & 0.7852\\
LG-ShadowNet (TIP'21)$\dagger$ & 23.0353 & 16.6409 & 17.9581 & 25.4526 & 19.6434 & 18.0556 & 0.9352 & 0.8352 & 0.7629\\
G2R-ShadowNet (CVPR'21+Mask)$\ddagger$ & 22.0473 & 18.6220 & 19.3276 & 24.6951 & 17.9899 & 16.8325 & 0.9015 & 0.8189 & 0.7183\\
DMTN (TMM'23+Mask)$\P$ & 11.5195 & 9.9282 & 10.2560 & 31.6098 & 24.1577 & 23.0287 & 0.9664 & 0.8696 & 0.8195\\
ST-CGAN (CVPR'18+Mask)$\P$ & 15.0976 & 13.0098 & 13.4399 & 27.4801 & 21.5584 & 20.2338 & 0.9525 & 0.8176 & 0.7486\\
DHAN (AAAI'20+Mask)$\P$ & \textbf{9.1524} & \textbf{9.0602} & \textbf{9.1086} & \textbf{33.0654} & \textbf{25.1760} & \textbf{24.1399} & \textbf{0.9752} & \textbf{0.9010} & \textbf{0.8620}\\
ShadowFormer (AAAI'23+Mask)$\P$ & \underline{9.6703} & \underline{9.2937} & \underline{9.5927} & \underline{32.7140} & \underline{24.3702} & \underline{23.4432} & \underline{0.9711} & \underline{0.8804} & \underline{0.8350}\\
TBRNet (TNNLS'23+Mask)$\P$ & 11.4115 & 10.8361 & 10.9546 & 31.2298 & 23.6601 & 22.5798 & 0.9619 & 0.8313 & 0.7722\\
S2-ShadowNet & 14.8059 & 12.9814 & 13.1006 & 30.7410 & 22.9772 & 22.0053 & 0.9570 & 0.8509 & 0.7881\\
\hline
\end{tabular}}
\caption{Quantitative comparisons on \textbf{UAV-SC}. Best and Second-Best scores are \textbf{highlighted} and \underline{underlined}. $\dagger$, $\ddagger$, and $\P$ represent unsupervised, weakly supervised, and fully supervised methods, respectively.}
\end{table*}
\subsection{Visual Comparisons}
We first show the comparisons on shadow images with complex cast-surface geometric shapes and boundaries sampled from WSSR in Figure \ref{fig5}.
All traditional-based methods fail to cope with shadow effects, \textit{e.g.}, Silva introduces color artifacts, Gong changes original tones, and Guo produces local degradation.
Although Mask-ShadowGAN, DC-ShadowNet, and LG-ShadowNet remove shadow traces to some extent, the consistency and coordination of shadow and shadow-free regions fail to be repaired.
In addition, G2R-ShadowNet introduces distorted illumination compensation, which seriously destroys the structure and pattern of remote sensing imagery.
In contrast, our method removes shadows while maintaining color consistency, which is credited to the fact that the infrared modality provides crucial illumination prior for the visible modality.

We then show the comparisons on challenging shadow images sampled from URSSR in Figure \ref{fig6}.
Silva introduces reddish and greenish color deviations.
Non-homogeneous shadow distribution challenges Gong and Guo.
This is because 1) accurate annotations are demanding for users, 2) inaccurate shadow detection exacerbates unsatisfactory performance.
G2R-ShadowNet fails to handle continuously varying shadow intensities, thus the robustness is not convincing.
Mask-ShadowGAN, DC-ShadowNet, and LG-ShadowNet are attractive for illumination recovery, but the texture details of the mountains are inevitably hidden.
In contrast, our method not only relights the dark pixels, but also maintains realistic and rich details.

We also show the comparisons on shadow images with illumination variations and shadow overlaps sampled from UAV-SC in Figure \ref{fig7}.
Cast shadow is tricky for Silva, Gong, and Guo because tiny shadow remnants are difficult to localize implicitly.
Competing unsupervised \cite{Hu:ICCV-19, Jin:ICCV-21, Liu:TIP-21} and weakly supervised \cite{Liu:CVPR-21} methods either render shadow remnants or introduce artifacts.
Besides, some fully supervised methods fail to maximally preserve detail information, such as DMTN and ST-CGAN.
In contrast, our method, DHAN, ShadowFormer, and TBRNet are closer to ground truths.

\begin{table}[!t]
\centering
\renewcommand\arraystretch{1.0}
\label{table2}
\setlength{\tabcolsep}{4pt}
\resizebox{!}{0.09\textwidth}{
\begin{tabular}{|c|c|c|}
\hline
Methods & VNM($\uparrow$) & Entropy($\uparrow$)\\
\hline
\hline
Silva (ISPRS'18) & 0.2461 & 6.5851\\
Gong (BMVC'14) & 0.2437 & 6.5510\\
Guo (TPAMI'13) & 0.2459 & 6.6002\\
Mask-ShadowGAN (ICCV'19)$\dagger$ & 0.2581 & 6.7909\\
DC-ShadowNet (ICCV'21)$\dagger$ & 0.2572 & 6.7123\\
LG-ShadowNet (TIP'21)$\dagger$ & \underline{0.2682} & \underline{6.8737}\\
G2R-ShadowNet (CVPR'21+Mask)$\ddagger$ & 0.2519 & 6.6778\\
S2-ShadowNet & \textbf{0.3062} & \textbf{6.9418}\\
\hline
\end{tabular}}
\caption{Quantitative comparisons on \textbf{WSSR}.}
\end{table}
\begin{table}[!t]
\centering
\renewcommand\arraystretch{1.0}
\label{table3}
\setlength{\tabcolsep}{4pt}
\resizebox{!}{0.09\textwidth}{
\begin{tabular}{|c|c|c|}
\hline
Methods & VNM($\uparrow$) & Entropy($\uparrow$)\\
\hline
\hline
Silva (ISPRS'18) & 0.2527 & 6.5902\\
Gong (BMVC'14) & 0.2540 & 6.6077\\
Guo (TPAMI'13) & 0.2659 & 6.6216\\
Mask-ShadowGAN (ICCV'19)$\dagger$ & 0.2926 & 6.6470\\
DC-ShadowNet (ICCV'21)$\dagger$ & 0.2966 & 6.6160\\
LG-ShadowNet (TIP'21)$\dagger$ & \underline{0.3138} & \underline{6.7738}\\
G2R-ShadowNet (CVPR'21+Mask)$\ddagger$ & 0.2868 & 6.5983\\
S2-ShadowNet & \textbf{0.3582} & \textbf{7.0356}\\
\hline
\end{tabular}}
\caption{Quantitative comparisons on \textbf{URSSR}.}
\end{table}
\subsection{Quantitative Comparisons}
For fair quantitative comparisons, we employ the source code provided by the authors, then retrain the compared methods using the consistent training set and achieve the best quantitative scores.
We first report the average RMSE, PSNR, and SSIM scores of the developed and compared methods in Table 1.
Our method outperforms traditional-based, unsupervised, and weakly supervised methods on the UAV-SC benchmark.
Admittedly, the fully supervised method has overwhelming superiority over the weakly supervised method in full-reference evaluations.
Our method is comparable to some fully supervised methods.
Coupled with the rigorous and expensive data acquisition of the fully supervised paradigm, our method is encouraging.

We then report the average VNM and Entropy scores on WSSR and URSSR benchmarks in Tables 2 and 3.
Compared with the top-performing LG-ShadowNet, our method achieves the percentage gain of 14.2\%/0.9\% and 14.1\%/3.9\% in terms of VNM/Entropy on WSSR and URSSR, respectively.
Such quantitative scores demonstrate the superiority of our method for shadow removal.

\begin{table}[!t]
\centering
\setlength{\tabcolsep}{10pt}
\resizebox{!}{0.088\textwidth}{
\begin{tabular}{|c|c|c|c|}
\hline
\multirow{2}{*}{Baselines} & \multicolumn{3}{c|}{UAV-SC}\\
\cline{2-4}
& RMSE($\downarrow$) & PSNR($\uparrow$) & SSIM($\uparrow$)\\
\hline
\hline
w/o MT & 19.7614 & 16.8275 & 0.7212\\
w/o ST & 16.5299 & 20.8440 & 0.7483\\
w/o $\mathcal{L}_{ort}$ & 15.3131 & 20.9069 & 0.7608\\
w/o $\mathcal{L}_{sim}$ & 14.8689 & 21.0520 & 0.7768\\
w/o $\mathcal{L}_{adv}$ & 19.9431 & 16.7428 & 0.7341\\
w/o $\mathcal{L}_{ide}$ & 17.3769 & 19.8943 & 0.7391\\
full model & \textbf{13.1006} & \textbf{22.0053} & \textbf{0.7881}\\
\hline
\end{tabular}
}
\caption{Quantitative scores of the ablation study in terms of RMSE, PSNR, and SSIM scores.\label{table4}}
\end{table}
\subsection{Ablation Studies}
We conduct extensive ablation studies to analyze the core components of S2-ShadowNet, including the modal translation and the spherical transform.
In addition, we analyze the linear combination of the orthogonality loss $\mathcal{L}_{ort}$, the similarity loss $\mathcal{L}_{sim}$, the adversarial loss $\mathcal{L}_{adv}$, and the identical loss $\mathcal{L}_{ide}$.
More specifically,
\begin{itemize}
\item w/o MT refers to S2-ShadowNet without the modal translation, similar to \cite{Liu:CVPR-21}, employing shadow and random shadow-free regions to accomplish weakly supervised learning.

\item w/o ST refers to S2-ShadowNet without the spherical transform, instead employing feature-wise concatenation to integrate visible and infrared representations.

\item w/o $\mathcal{L}_{ort}$ means that S2-ShadowNet is trained without the constraint of the orthogonality loss.

\item w/o $\mathcal{L}_{sim}$ means that S2-ShadowNet is trained without the constraint of the similarity loss.

\item w/o $\mathcal{L}_{adv}$ means that S2-ShadowNet is trained without the constraint of the adversarial loss.

\item w/o $\mathcal{L}_{ide}$ means that S2-ShadowNet is trained without the constraint of the identical loss.
\end{itemize}

\begin{figure}[t]
\centering
\includegraphics[width=0.95\columnwidth]{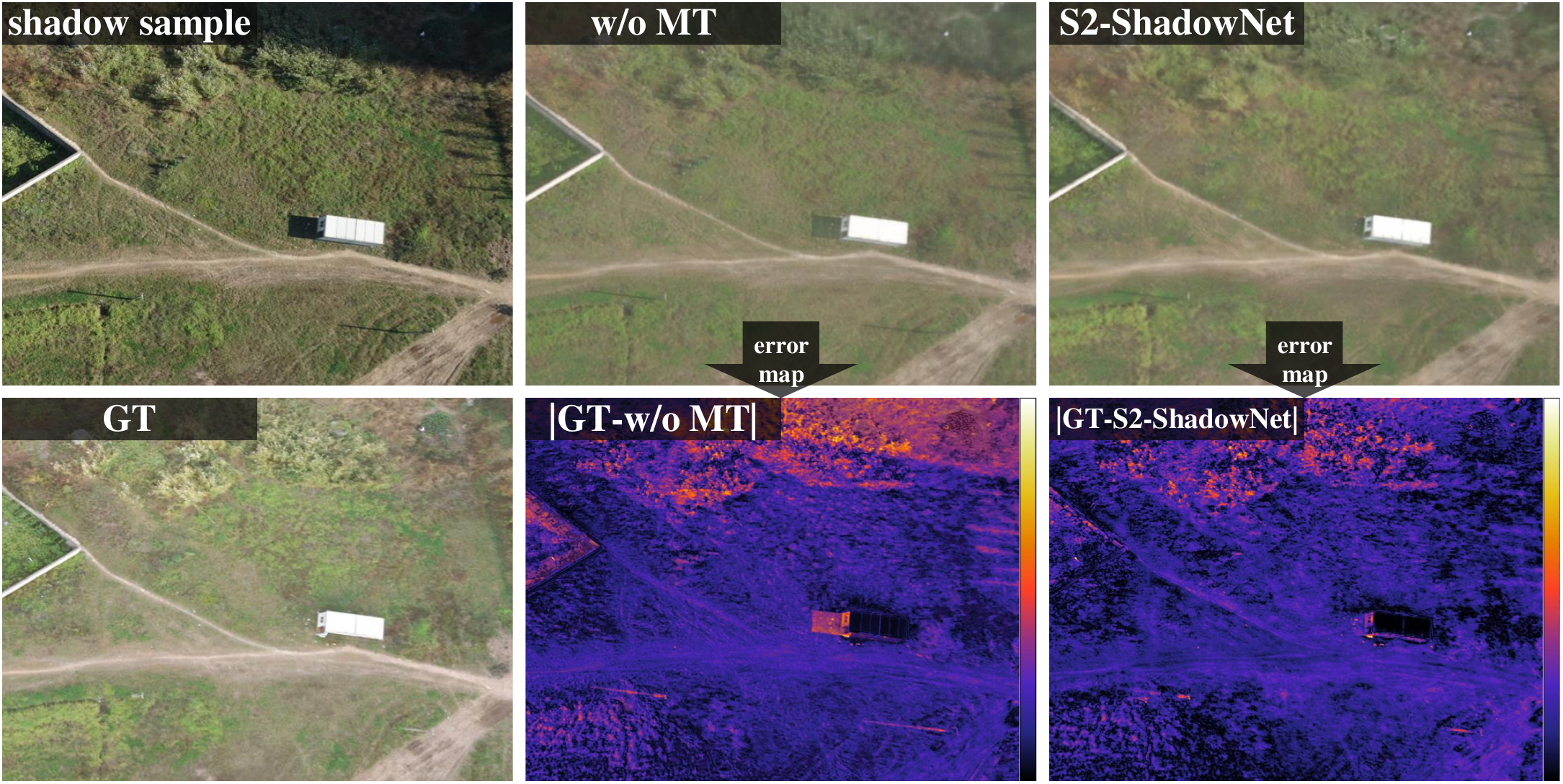}
\caption{Ablation study towards the modal translation. Error maps suggest that the introduction of infrared modality contributes to remove shadow traces.}
\label{fig8}
\end{figure}
\begin{figure}[t]
\centering
\includegraphics[width=0.95\columnwidth]{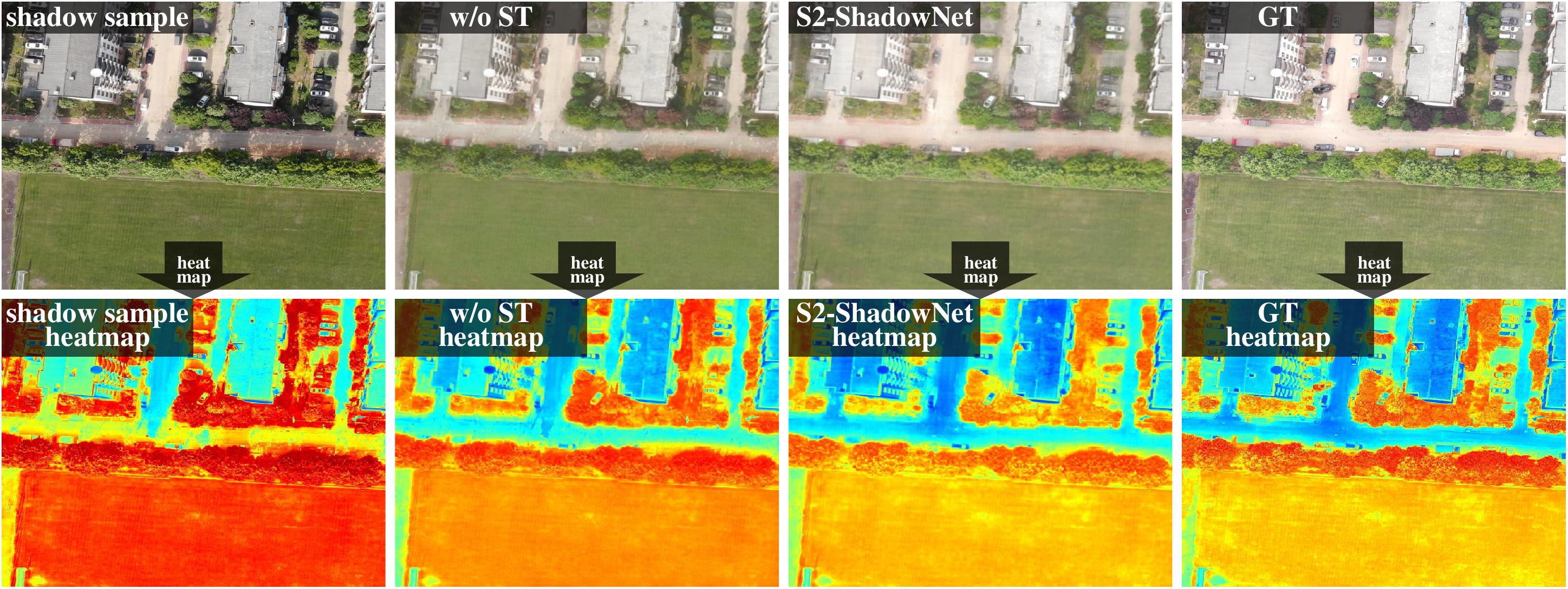}
\caption{Ablation study towards the spherical transform. Heatmaps suggest that the aggregation of multiple modalities in spherical space achieves a double win across illumination compensation and detail restoration.}
\label{fig9}
\end{figure}
\begin{figure}[t]
\centering
\includegraphics[width=0.95\columnwidth]{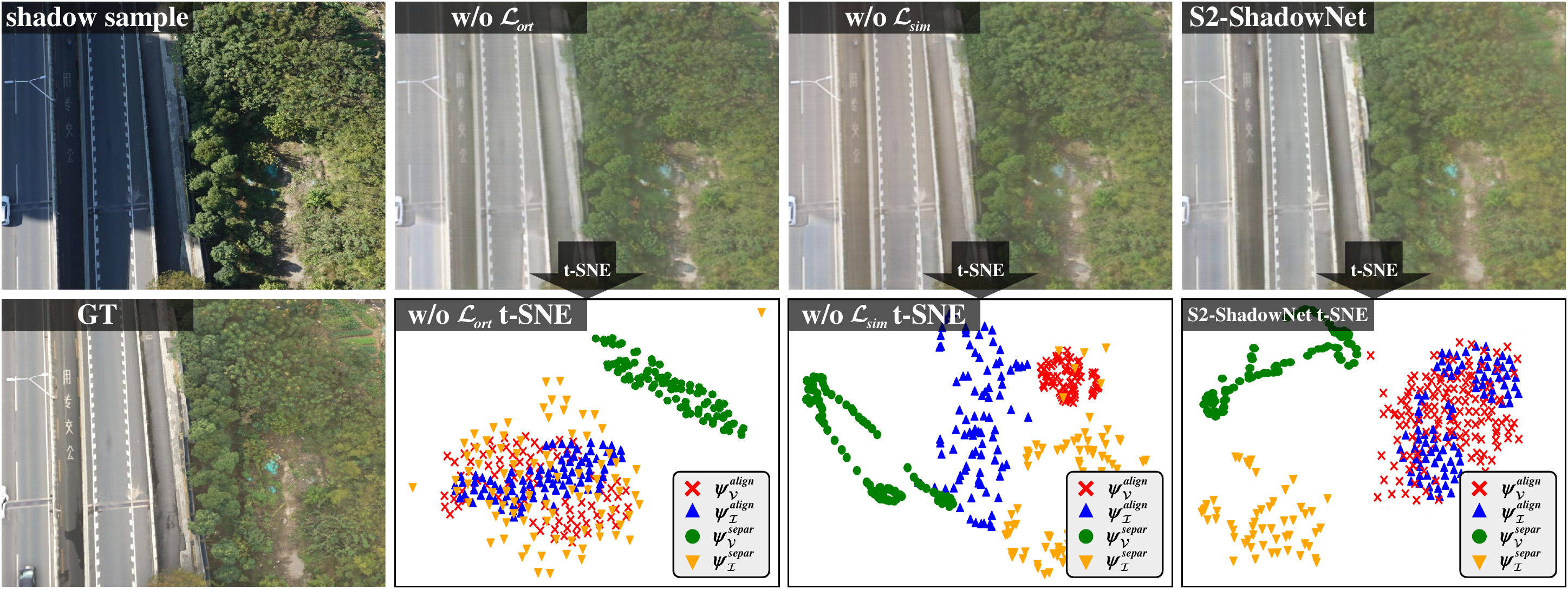}
\caption{Ablation study towards the orthogonality loss and the similarity loss. t-SNE suggests that the collaboration between orthogonality loss and similarity loss gracefully accomplishes cross-modal feature decomposition.}
\label{fig10}
\end{figure}
\begin{figure}[t]
\centering
\includegraphics[width=0.95\columnwidth]{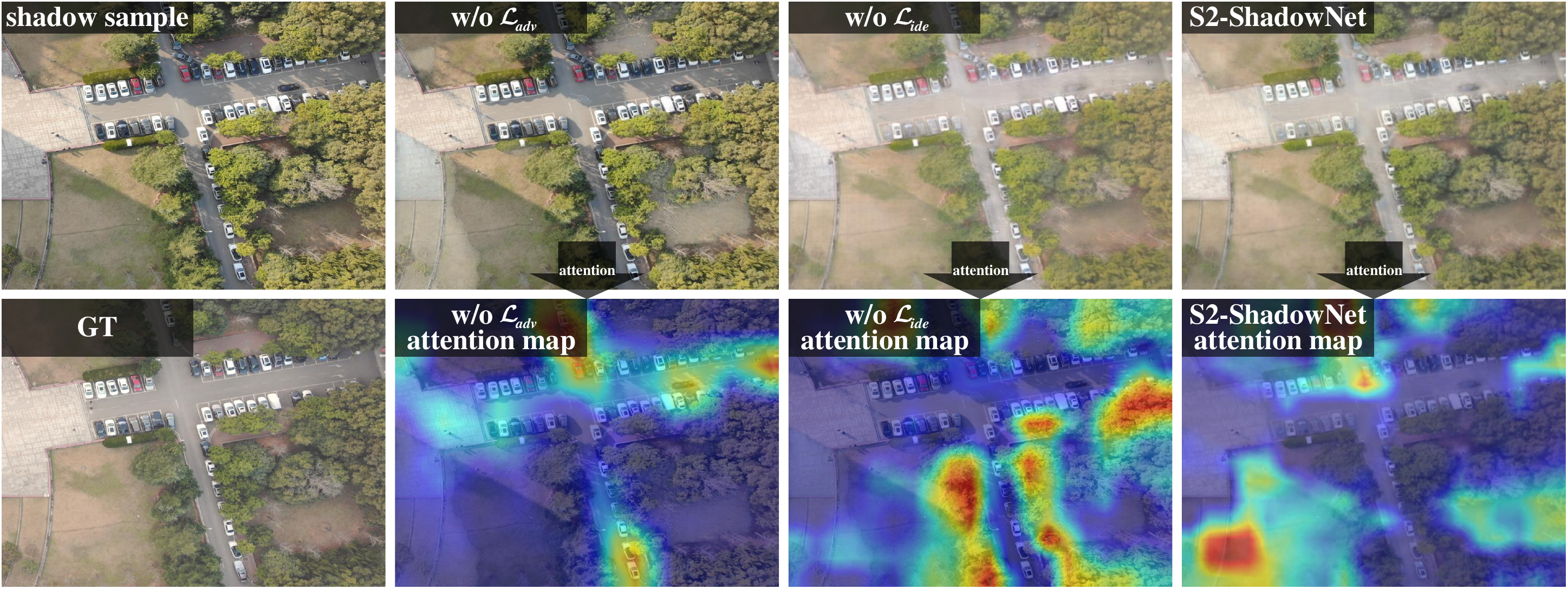}
\caption{Ablation study towards the adversarial loss and the identical loss. Feature visualizations suggest that the interaction between adversarial loss and identical loss focuses more attention on shadow remnants.}
\label{fig11}
\end{figure}

The quantitative scores of the ablated models on UAV-SC are reported in Table 4.
In addition, the contributions of modal translation, the effectiveness of spherical transform, and the effects of loss function are depicted in Figs. \ref{fig8}, \ref{fig9}, \ref{fig10}, and \ref{fig11}, respectively.
The conclusions drawn from the ablation study are listed as follows.
\begin{itemize}
\item As shown in Table 4, the full model achieves the best RMSE, PSNR, and SSIM scores, which implies that the introduction of infrared modality and the design of spherical space transform are convincing.

\item In Fig. \ref{fig8}, the ablated model w/o MT produces undesired shadow traces, which are obvious in error visualizations.
In contrast, the full model removes shadow boundaries through the close cooperation of visible and infrared information.

\item Arbitrary and diverse cast shadows challenge the ablated model w/o ST, as shown in Fig. \ref{fig9}.
The heatmap indicates that aimlessly integrating multi-modal information fails to recover promising illumination.
In contrast, the heatmap of the full model is closer to ground truth in terms of detail and texture.
Therefore, employing spherical space to perform domain aggregation is imperative.

\item As shown in Fig. \ref{fig10}, the ablated models w/o $\mathcal{L}_{ort}$ and w/o $\mathcal{L}_{sim}$ either fail to separate private and shared features or fail to align shared features.
In contrast, the full model integrates multi-modal shared features while pushing them away from multi-modal private features for attractive cross-modal modeling.

\item In Fig. \ref{fig11}, we observe obvious shadow effects without employing the adversarial loss $\mathcal{L}_{adv}$.
This is because the lack of $\mathcal{L}_{adv}$ fails to provide shadow generation that supports weakly supervised learning.
Besides, the ablated model w/o $\mathcal{L}_{ide}$ fails to focus attention on shadow regions, leading to shadow remnants.
The full model allocates more shadow-aware attention to quality-degraded regions, thus achieving visually pleasing quality.
\end{itemize}


\section{Conclusion}
In this work, we propose a weakly supervised shadow removal model.
The core insight of S2-ShadowNet is employing spherical space to accomplish the alignment/separation of domain-shared/private representations, thus removing highly-complex shadow effects.
In addition, we contribute a high definition shadow removal benchmark that makes weakly supervised shadow removal out of the constraints of specific scenarios possible.
Extensive experiments have demonstrated the superiority of S2-ShadowNet.

\bibliography{aaai25}

\end{document}